\documentclass[sigconf]{acmart}
\def\BibTeX{{\rm B\kern-.05em{\sc i\kern-.025em b}\kern-.08emT\kern-.1667em\lower.7ex\hbox{E}\kern-.125emX}}
    
\renewcommand\footnotetextcopyrightpermission[1]{} 
\usepackage{verbatim}
\usepackage{amsmath}
\usepackage{amssymb}
\usepackage{booktabs} 
\usepackage{algorithm}
\usepackage{algorithmicx} 
\usepackage{algcompatible}
\usepackage{algpseudocode}
\usepackage{multirow}
\usepackage{color}
\usepackage{threeparttable}
\usepackage{pifont}
\usepackage{etaremune}
\usepackage{caption}
\usepackage{subcaption}
\usepackage[export]{adjustbox}

\usepackage{enumitem}
\setlist{leftmargin=3.0mm}

\usepackage{soul}

\usepackage{tikz}
\usepackage{calc}

\begin{document}
\pagestyle{plain}
\pagenumbering{gobble}

\title{
\vspace{-26pt} EDD: Efficient Differentiable DNN Architecture and Implementation Co-search for Embedded AI Solutions \vspace{-8pt}
}

\affiliation{\large{ 
Yuhong Li$^{1*}$, 
Cong Hao$^{1*}$, 
Xiaofan Zhang$^{1}$,
Xinheng Liu$^{1}$,
Yao Chen$^{2}$,
Jinjun Xiong$^{3}$,
Wen-mei Hwu$^{1}$,
Deming Chen$^{1}$}
}

\thanks{$*$These authors made equal contributions.}

\affiliation{ \normalsize \vspace{-10pt}
	\institution{$^1$Electrical and Computer Engineering, University of Illinois at Urbana-Champaign\\ $^2$Advanced Digital Sciences Center, Singapore, $^3$IBM T. J. Watson Research Center}
}
\affiliation{ \normalsize
	\textit{\{leeyh, congh, xiaofan3, xliu79, w-hwu, dchen\}@illinois.edu, yao.chen@adsc-create.edu.sg, jinjun@us.ibm.com}
}

\begin{abstract}
High quality AI solutions require joint optimization of AI algorithms and their hardware implementations.
In this work, we are the first to propose a fully simultaneous,
\textbf{E}fficient \textbf{D}ifferentiable \textbf{D}NN (deep neural network) architecture and implementation co-search (\textbf{EDD}) methodology.
We formulate the co-search problem by fusing DNN search variables and hardware implementation variables into one solution space, and maximize both algorithm accuracy and hardware implementation quality.
The formulation is differentiable with respect to the fused variables, so that gradient descent algorithm can be applied to greatly reduce the search time.
The formulation is also applicable for various devices with different objectives.
In the experiments, we demonstrate the effectiveness of our EDD methodology by searching for three representative DNNs, targeting low-latency GPU implementation and FPGA implementations with both recursive and pipelined architectures. Each model produced by EDD achieves similar accuracy as the best existing DNN models searched by neural architecture search (NAS) methods on ImageNet, but with superior performance obtained within 12 GPU-hour searches.
Our DNN targeting GPU is $1.40\times$ faster than the state-of-the-art solution reported in Proxyless~\cite{cai2018proxylessnas}, and our DNN targeting FPGA delivers $1.45\times$ higher throughput than the state-of-the-art solution reported in DNNBuilder~\cite{zhang2018dnnbuilder}.

\end{abstract}

\maketitle

\vspace{-6pt}
\section{Introduction}

AI algorithms have gained ever-increasing research interests, and
remarkable achievements have been demonstrated especially for deep neural networks (DNNs).
To improve algorithm quality, usually accuracy, a recent approach called neural architecture search (NAS) \cite{zoph2017learning, huang2017densely, real2018regularized} has shown great success in automatically developing DNNs that outperform human-crafted designs.
Meanwhile, the optimization techniques of implementing AI algorithms on hardware are also being intensively studied.
The goal of implementation is to improve hardware performance, such as latency, throughput and energy efficiency, etc.
Typical implementation techniques include kernel and DNN optimizations on GPUs and customized accelerator designs on FPGAs and ASICs~\cite{qiu2016going,zhang2018dnnbuilder,x2017High,chen2019cloud, hao2019fpga, xu2020autodnnchip}.
On top of these achievements, to further improve AI solution quality, AI algorithm designers and hardware developers begin to explore joint optimization opportunities.
For example, hardware-aware NAS has been drawing a lot of attention by considering hardware features for DNN designs \cite{cai2018proxylessnas, tan2019mnasnet, stamoulis2019single, wu2019fbnet, gong2019mixed, zhang2020skynet}.
Meanwhile, hardware/software co-design approaches ~\cite{jiang2019accuracy, hao2019fpga} focus on FPGA implementation characteristics and study their influences on DNN software design.

Despite these achievements of hardware-aware NAS and hardware/software co-design,
there is still a large optimization opportunity missing:
\textit{the hardware implementation should be simultaneously searched during NAS}.
Here, for general-purpose computing devices, such as CPUs and GPUs, \textit{implementation search} means optimizing DNN implementations such as kernel fusion and memory access optimization.
For reconfigurable devices, such as FPGAs, \textit{implementation search} means optimizing a customized DNN accelerator through techniques such as quantization, loop tiling and parallelization.
Hardware implementation search not only provides more accurate performance evaluations on latency and throughput, but more importantly, provides instant guidance to hardware-aware DNN design during NAS.
Currently, all existing works are missing the large design space of \textit{implementation search} in their NAS flows, using estimated hardware performance from a fixed implementation~\cite{cai2018proxylessnas, tan2019mnasnet, stamoulis2019single, wu2019fbnet, gong2019mixed}.
We provided an initial discussion of the potential of simultaneous neural architecture and implementation co-search in~\cite{hao2019nais}, called NAIS.
Inspired by~\cite{hao2019nais}, in this work, we propose a simultaneous and efficient DNN and implementation co-search methodology.
We summarize our contributions as follows:

\begin{itemize}
 \item {
    This is the first work that proposes a mathematical formulation to solve the simultaneous DNN architecture and hardware implementation co-search problem. The formulation fuses the search variables for DNN architecture and its hardware implementation into one search space, to simultaneously maximize the DNN accuracy and implementation performance.}
    \item {
    The proposed formulation is differentiable with respect to the fused search space, which allows a gradient descent algorithm to be applied and enables an efficient differentiable DNN and implementation co-search methodology (EDD).
    }
    \item{
    The formulation is unified and comprehensive. It can be applied on various hardware platforms such as GPUs, FPGAs and dedicated accelerators; it can target various performance objectives such as latency, throughput or energy; it also formulates resource usage considering resource sharing. 
    }
    
    \item {
    We demonstrate our EDD methodology targeting three different architectures: GPU, recursive FPGA accelerator, and pipelined FPGA accelerator. Each model produced by EDD achieves similar accuracy as the best existing DNNs on ImageNet but with superior performance: our GPU-targeted DNN is $1.40\times$ faster than the state-of-the-art Proxyless solution~\cite{cai2018proxylessnas}, and our FPGA-targeted DNN delivers $1.45\times$ higher throughput than the state-of-the-art DNNBuilder solution~\cite{zhang2018dnnbuilder}.
    
    }
    
\end{itemize}


\vspace{-8pt}
\section{Related Works}
\label{sec:related_work}
Neural architecture search (NAS) is a technique for automating the design of DNNs which can outperform hand-crafted ones~\cite{zoph2017learning, huang2017densely, real2018regularized}.
Among all the NAS approaches,
differentiable NAS is becoming prevailing because of its high search efficiency in terms of GPU hours \cite{liu2018darts}. 
One typical way is to construct a DNN supernet composed of multiple branch candidates associated with architecture weights. The architecture weights are differentiable with respect to the loss function of the supernet, and are updated through stochastic gradient descent.
The final searched DNNs will keep the branches with larger architecture weights and eliminate others.
For example, \cite{cai2018proxylessnas} uses binarized parameters, and \cite{wu2019fbnet} uses Gumbel-Softmax to choose between different DNN branches.
Previously published literature also introduces hardware-aware NAS~\cite{cai2018proxylessnas, tan2019mnasnet, stamoulis2019single, wu2019fbnet, gong2019mixed}.
Some works incorporate hardware latency into the objective of NAS~\cite{cai2018proxylessnas, stamoulis2019single, wu2019fbnet}, while some treat latency as a hard constraint~\cite{tan2019mnasnet}.
In \cite{zhang2020skynet}, a bottom-up DNN design approach is proposed for hardware-efficient models.


Meanwhile, various embedded FPGA-based accelerators have been studied to support more efficient DNN inference \cite{qiu2016going, zhang2018dnnbuilder}. 
Other approaches proposed in \cite{jiang2019accuracy, hao2019fpga} involve hardware/software co-design.
Specifically, researchers in~\cite{jiang2019accuracy} proposed a reinforcement learning based architecture search with FPGA implementation performance integrated into the reward function.
The work in~\cite{hao2019fpga} proposed a bundle-based co-design methodology, where a bundle is the basic building block for both FPGA accelerator and DNN model.
However, none of the previous works is able to explore the DNN architecture and implementation co-search space simultaneously and comprehensively.

\vspace{-4pt}
\section{EDD problem formulation \label{sec:nais_design_space}}

\begin{figure*}
\vspace{-18pt}
    \centering
    \includegraphics[width=0.91\textwidth]{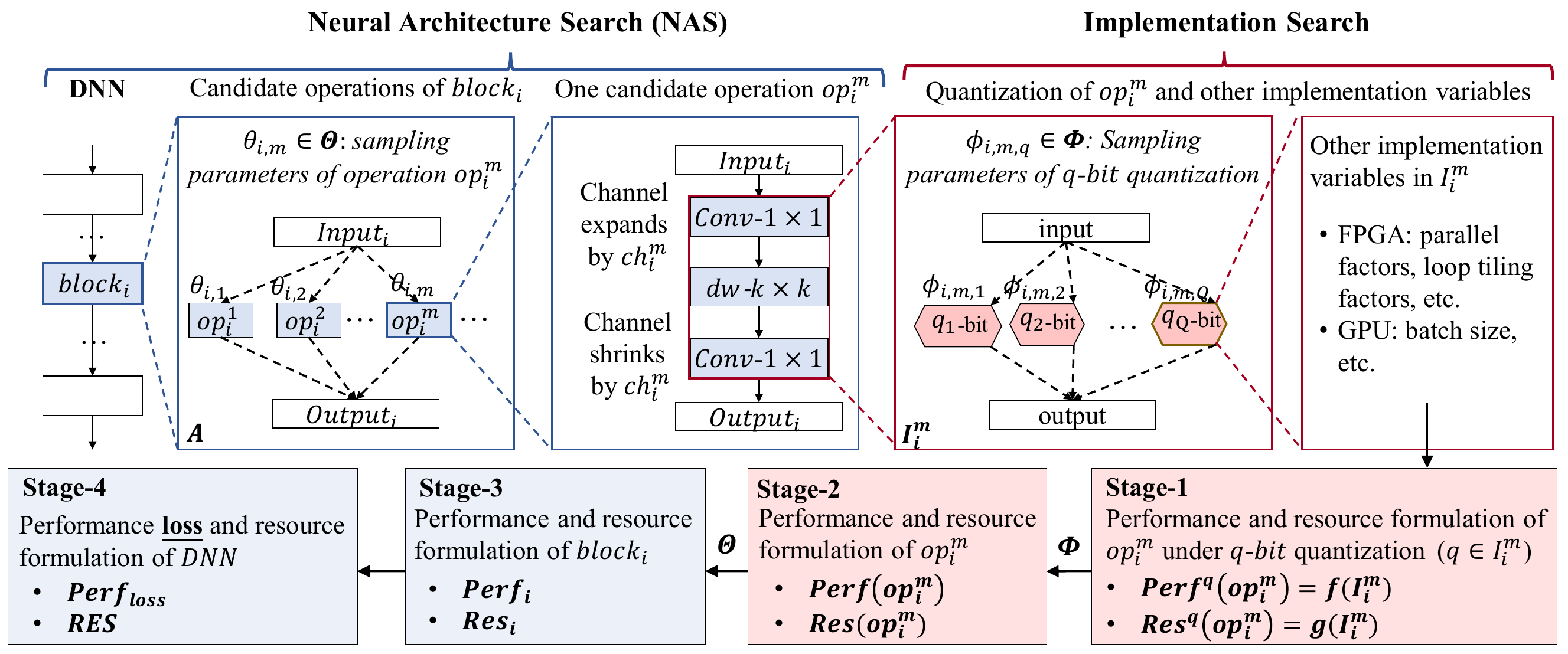}
    \vspace{-8pt}
    \caption{Our proposed differentiable DNN and implementation co-design (EDD) space.}
    \label{fig:nais_design_space}
    \vspace{-4pt}
\end{figure*}

Our simultaneous DNN and implementation co-search method fuses the design space of DNN architecture search and hardware implementation search, as shown in Fig.~\ref{fig:nais_design_space}.
We collectively denote the variables used in DNN search and implementation search as $A$ and $I$, respectively, and the fused space of co-search is $\{A, I\}$.
The objective of DNN search is to quickly find a DNN architecture while minimizing accuracy loss, denoted as $Acc_{loss}$.
For implementation search, we define \textit{performance loss}, denoted as $Perf_{loss}$, which can be specified by users, such as end-to-end inference latency, throughput, energy, DNN model complexity, etc. We denote the resource utilization as $RES$, and the resource upper-bound of the target hardware as $RES_{ub}$.
The DNN and implementation co-search problem is to minimize accuracy loss $Acc_{loss}$ and performance loss $Perf_{loss}$ simultaneously by effectively searching $\{A, I\}$:
\begin{equation}
    min:\mathcal{L} = Acc_{loss}(A,I) \cdot Perf_{loss}(I) + \beta \cdot C^{RES(I) - RES_{ub}}
    \label{eq:co_search_obj_func}
    \vspace{-4pt}
\end{equation}

In Eq.~\ref{eq:co_search_obj_func}, $Acc_{loss}$ is a function of $A$ and $I$;
$Perf_{loss}$ and $RES$ are functions of $I$.
Resource upper-bound $RES_{ub}$ is expressed in an exponent term to introduce large penalty when being violated.
Worth noting, in the existing hardware-aware NAS approaches, only $A$ is searched while $I$ is \textit{fixed} during NAS;
while in our co-search formulation, $I$ is also \textit{variable}.

As introduced in Sec. \ref{sec:related_work}, motivated by the high search efficiency and appealing model accuracy of differentiable NAS,
in this work, we propose a differentiable formulation for both $\{A, I\}$: in Eq.~\ref{eq:co_search_obj_func},
$Acc_{loss}$ is differentiable with respect to $A$ and $I$, and
$Perf_{loss}$ and $RES(I)$ are differentiable with respect to $I$.
By descending $\mathcal{L}$ with respect to the variables in $\{A, I\}$ on validation set as $\bigtriangledown_{\{A, I\}}\mathcal{L}_{val}$, $\{A,I\}$ will be searched simultaneously.


Fig.~\ref{fig:nais_design_space} shows our proposed overall differentiable design space.
The blue blocks represent the DNN search space, while the red blocks represent the hardware implementation search space.
We first introduce DNN search space in Sec.~\ref{sec:nas_design_space},
and then introduce the merged design space with implementation in Sec.~\ref{sec:impl_formulation}.

\vspace{-4pt}
\subsection{NAS Design Space\label{sec:nas_design_space}}

The differentiable NAS space is shown as the blue blocks in ~Fig.~\ref{fig:nais_design_space}.
First, the DNN is composed of $N$ basic building blocks, $block_i$, where $1\leq i \leq N$.
In this work, in order to design hardware-friendly DNNs and to reduce search time, we adopt the 
single-path DNN structure without branches \cite{stamoulis2019single}.
Inside the $i$-$th$ block, there are $M$ candidate operations, denoted as $op_i^m$ ($1 \leq m \leq M$).
We let the operations to be the most commonly used DNN blocks in NAS approaches, called MBConv~\cite{tan2019mnasnet}. It is composed of sequential layers of $conv$-$1\times1$, $dwconv$-$k\times k$ (depth-wise convolution with kernel size $k$) and $conv$-$1\times1$.
Between $conv$-$1\times1$ and $dwconv$-$k\times k$, the number of channels expands/shrinks by a ratio of $ch_i^m$.
The output of a block is calculated based on the outputs of its $M$ candidate operations. For example in \cite{liu2018darts}, the output is the weighted sum of the $M$ operations, where the weights are determined by a Softmax function.
Instead of Softmax, in this work, we use the Gumbel-Softmax function in \cite{wu2019fbnet} in order to sample
only one operation out of $M$ during feedforward propagation,
since Gumbel-Softmax function can convert the discrete non-differentiable sampling to continuous differentiable sampling.
This greatly reduces the memory requirement and speeds up the feedforward propagation.
The sampling parameters $\theta_{i,m}$ organize a two-dimension $N\times M$ array, denoted as $\Theta \in A$, which is the primary DNN search variable.

\vspace{-6pt}
\subsection{Implementation Formulation \label{sec:impl_formulation}}

As shown in the red block in Fig.~\ref{fig:nais_design_space},
each candidate operation $op_i^m$ has its own implementation variables, forming an implementation search space $I_i^m$.
The primary implementation variable is quantization $q$, i.e., data precision, since it has a large impact on DNN accuracy, implementation performance and hardware resource.
Rather than a train-and-quantize manner, the quantization shall be searched together with DNN structure to provide implementation performance feedback.
Besides quantization, other implementation variables may be device oriented. For example, FPGA implementation design space includes parallelism, loop tiling factors, etc.

To formulate the final $Perf_{loss}$ and $RES$ of a DNN in Eq.~\ref{eq:co_search_obj_func},
we need to capture the intermediate performance and resource of each operation and DNN block.
As shown in the bottom four blocks in Fig.~\ref{fig:nais_design_space},
there are four stages to derive Eq.~\ref{eq:co_search_obj_func}:
\vspace{-4pt}
\begin{itemize}
    \item {
    Stage-1: we first formulate the quantization in a differentiable way; then, for each operation candidate $op_i^m$ under $q$-bit, we formulate the performance as $Perf^q(op_i^m)$ and resource as $Res^q(op_i^m)$.
    }
    \item {
    Stage-2: the performance and resource of $op_i^m$ regardless of quantization, $Perf(op_i^m)$ and $Res(op_i^m)$, can be derived from $Perf^q(op_i^m)$ and $Res^q(op_i^m)$ in Stage-1.
    }
    \item {
    Stage-3: the performance and resource of $i$-$th$ DNN block, $Perf_i$ and $Res_i$, are derived from $Perf(op_i^m)$ and $Res(op_i^m)$ in Stage-2.
    }
    \item {
    Stage-4: the overall DNN performance loss and resource usage, $Perf_{loss}$ and $RES$, are derived from $Perf_i$ and $Res_i$ in Stage-3.
    }
    \item {
    Finally, $Perf_{loss}$ and $RES$ will be plugged into Eq.~\ref{eq:co_search_obj_func} as the objective function during our EDD co-search.
    }
\end{itemize}

In the following, we introduce the differentiable quantization and performance and resource formulations stage by stage.

\vspace{2pt}
\textbf{3.2.1 Stage-1: Differentiable Quantization}
\vspace{2pt}

As shown in the red block in Fig.~\ref{fig:nais_design_space}, to enable differentiable quantization formulation, we create $Q$ quantization paths for each operation $op_i^m$, indicating each operation has $Q$ quantization choices, from $q_1$-$bit$ to $q_Q$-$bit$.
Similar to differentiable NAS formulation, each quantization scheme is also sampled by the Gumbel-Softmax function with a sampling parameter $\phi_{i, m, q}$, generating a possibility for $op_i^m$ to be quantized to $q$-$bit$.
The $\phi_{i, m, q}$ organizes a three-dimension array of size $N\times M\times Q$, denoted as $\Phi$.
In this formulation, we have the flexibility to choose different quantizations for different layers of a DNN; such a mixed precision computation can be well supported by reconfigurable hardware and dedicated accelerators. 

Under $q$-$bit$ quantization, the performance and resource of operation $op_i^m$, $Perf^q(op_i^m)$ and $Res^q(op_i^m)$,
should be functions of implementation variables in $I_i^m$ (including quantization $q$), expressed as $Perf^q(op_i^m) = f(I_i^m)$ and $Res^q(op_i^m) = g(I_i^m)$.
Since one operation contains multiple layers as shown in Fig.~\ref{fig:nais_design_space}, for simplicity, we treat them as a whole: the $q$-$bit$ quantization applies to all layers within $op_i^m$, and the latency and resource are the summation of all layers.
$Perf^q(op_i^m)$ and $Res^q(op_i^m)$ largely vary with devices and will be further discussed in Section \ref{sec:device-specific}.

\begin{figure}[t]
    \centering
    \includegraphics[width=0.42\textwidth]{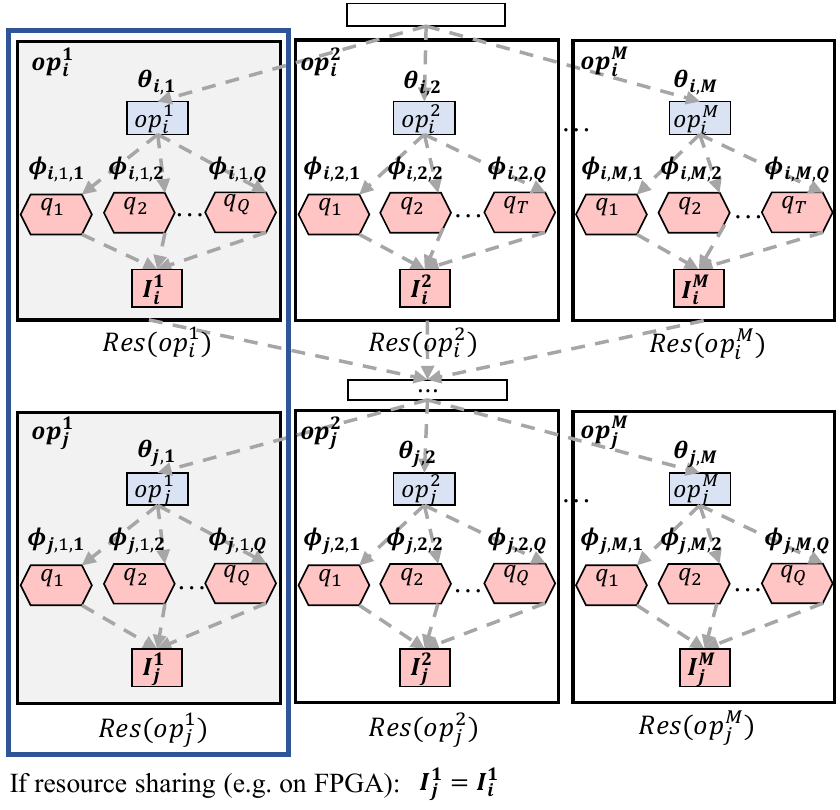}
    \vspace{-8pt}
    \caption{Fused design space including $\Theta$ (for DNN), $\Phi$ (for quantization) and other implementation variables in $I$.}
    \label{fig:merged-design-space}
    \vspace{-12pt}
\end{figure}

Given differentiable DNN search variables $\Theta$, differentiable quantization variables $\Phi$ and formulations under each quantization scheme, the DNN and implementation search spaces are fused as shown in Fig.~\ref{fig:merged-design-space}.

\vspace{2pt}
\textbf{3.2.2 From Stage-1 to Stage-2}
\vspace{2pt}

Given array $\Phi$, following Gumbel-Softmax sampling rule, denoted as $GS(\cdot)$,
the performance $Perf(op_i^m)$ and resource $Res(op_i^m)$ can be computed as the following:
\begin{equation}
    Perf(op_i^m) = \sum_{1 \leq q\leq Q} GS( \phi_{i, m, q} | \phi_{i, m}) \cdot Perf^q(op_i^m)
    \label{eq:imp_op_i_m}
    \vspace{-2pt}
\end{equation}
\vspace{-2pt}
\begin{equation}
    Res(op_i^m) = \sum_{1 \leq q\leq Q} GS( \phi_{i, m, q} | \phi_{i, m}) \cdot Res^q(op_i^m)
    \label{eq:res_op_i_m}
    \vspace{-2pt}
\end{equation}
where both $Perf(op_i^m)$ and $Res(op_i^m)$ are differentiable with respect to $\phi_{i, m, q}$.
This is to compute the performance and resource expectation under different quantizations, which follows Gunbel-Softmax distribution with parameter $\phi_{i, m, q}$.

\vspace{2pt}
\textbf{3.2.3 From Stage-2 to Stage-3}
\vspace{2pt}

Similar to Eq.~\ref{eq:imp_op_i_m} and Eq.~\ref{eq:res_op_i_m}, given array $\Theta$, the performance and resource of $i$-th DNN block, can be expressed as:
\begin{equation}
    Perf_i = \sum_{1 \leq m \leq M} GS( \theta_{i, m} | \theta_{i}) \cdot Perf(op_i^m)
    \label{eq:per_i}
\end{equation}
\vspace{-4pt}
\begin{equation}
    Res_i = \sum_{1 \leq m \leq M} GS( \theta_{i, m} | \theta_{i}) \cdot Res(op_i^m)
    \label{eq:res_i}
\end{equation}

\vspace{2pt}
\textbf{3.2.4 From Stage-3 to Stage-4 --- Performance}
\vspace{2pt}

Given the performance and resource of $i$-th DNN block, we can compute the overall DNN performance loss and resource usage, which need to be tailored towards specific search objectives and different devices.

First, if the overall objective is end-to-end latency, total energy or model size, the performance loss can be expressed using the summation of all DNN blocks as:
\begin{equation}
    Perf_{loss} =  \alpha \cdot \sum_{1 \leq i \leq N} Perf_i
    \label{eq:perf_loss_latency}
\end{equation}
where $\alpha$ scales $Perf_{loss}$ to the same magnitude of $Acc_{loss}$ in Eq.~\ref{eq:co_search_obj_func}.

If the objective is throughput, the performance loss is
the maximum latency of all blocks.
Since getting the maximum value is a non-differentiable operation, we use a smooth maximum, Log-Sum-Exp (LSE) function~\cite{polak2012optimization}, for differentiable approximation as:
\begin{equation}
    Perf_{loss} =  \alpha \cdot max\{ Perf_i \} 
      \approx  \alpha \cdot log \sum_{1 \leq i \leq N} e^{Perf_i}
    \label{eq:Lat_max}
        \vspace{-2pt}
\end{equation}

If there are multiple objectives such as minimizing both latency and energy, as long as the objectives are not conflicting, we can simply let $Perf_{loss}$ be the production of different objectives.

\vspace{2pt}
\textbf{3.2.5 From Stage-3 to Stage-4 --- Resource}
\vspace{2pt}

The formulation of overall resource usage has two situations: without and with resource sharing.
Without sharing, the total resource can be computed as the summation of all blocks:
\begin{equation}
    RES = \sum_{1 \leq i \leq N} Res_i
    \label{eq:res_throughput}
        \vspace{-2pt}
\end{equation}

However, resource sharing is very common especially in IP-based FPGA or ASIC accelerators.
Fig.~\ref{fig:merged-design-space} demonstrates a resource sharing scenario. In this example, we assume the operation $op_i^1$ of $i$-$th$ block, and operation $op_j^1$ of $j$-$th$ block, will share a same piece of computing resource. For example, in FPGA or ASIC, it is a reusable IP.
To allow sharing, the quantization and other implementation variables of $op_i^1$ and $op_j^1$ shall be the same\footnote{Some accelerators allow different bit-width operations to share resource. In these cases this constraint is not needed.}:
 $for~\forall i, j\in [1,N],$ we have $I_{i}^m = I_{j}^m = I^m$.

\begin{figure}[t]
\vspace{-8pt}
    \centering
    \includegraphics[width=0.45\textwidth]{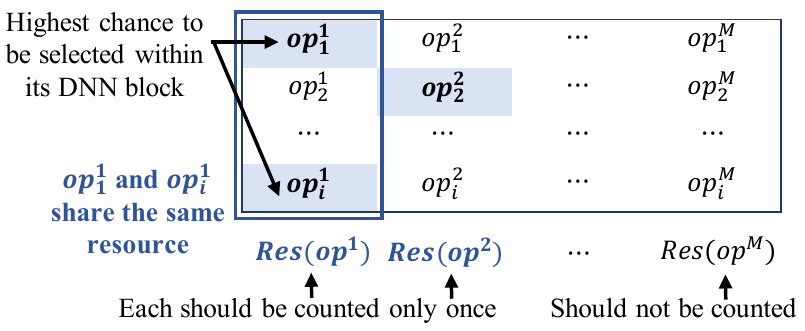}
    \vspace{-8pt}
    \caption{Demonstration of resource sharing.}
    \label{fig:merged-space-resource-sharing}
\vspace{-8pt}
\end{figure}

Second, we discuss the resource estimation with resource sharing.
As shown in Fig.~\ref{fig:merged-space-resource-sharing}, $i$-$th$ row is the $i$-$th$ DNN block; the blue entries are the operations with the largest possibility to be selected.
In this example, $op_1^1$ and $op_i^1$ are most likely to be chosen. Since they share the same computing resource $Res(op^1)$, it shall be counted only once.
If the $m$-$th$ operation is not selected in any of the blocks, the resource $Res(op^m)$ shall not be counted.

To describe resource sharing, we propose the following differentiable approximation for the resource usage for operation $op^m$, $Res(op^m)$, which is shared across blocks as:
\begin{equation}
    Res(op^m) \leftarrow tanh(\sum_{1\leq i \leq N} GS( \theta_{i, m} | \theta_{i}) \cdot 1 )\cdot Res(op^m)
    \label{eq:res_latency_1}
        \vspace{-2pt}
\end{equation}

In the above formula, 
$GS( \theta_{i, m} | \theta_{i}) \cdot 1$ means the unit resource expectation of $op_i^m$ in block $i$.
To avoid operation resource being redundantly counted across block, we use $tanh(\sum)$ to suppress the maximum expectation of $m$-th operation to be 1 before multiplied by $Res(op^m)$.

Thus, the overall DNN resource is computed as:
\begin{equation}
    RES = \sum_{1 \leq m \leq M} Res(op^m)
    \label{eq:res_latency_2}
        \vspace{-2pt}
\end{equation}

\vspace{-2pt}
\section{Device-specific formulation\label{sec:device-specific}}

In this section we discuss the device-specific formulations, which describe the performance and resource of operation $op_i^m$ under $q$-bit quantization, $Perf^q(op_i^m)$ and $Res^q(op_i^m)$.

\vspace{-4pt}
\subsection{FPGA}
For FPGA implementation, we follow an IP-based accelerator architecture:
for each operation $op_i^m$, there is a customizable IP instance to conduct its computation.
Consider the following two FPGA implementation architectures,
recursive and pipelined:
\begin{itemize}
    \item {
    In the recursive architecture such as \cite{chen2019cloud, hao2019fpga}, every DNN layer of the same type shares the same IP.
    The implementation objective is usually end-to-end latency.
    Thus, the $Perf_{loss}$ computation follows Eq.~\ref{eq:perf_loss_latency}, and the
    $RES$ follows Eq.~\ref{eq:res_latency_1} and Eq.~\ref{eq:res_latency_2}.
    }
    \item {
    In the pipelined architecture such as \cite{zhang2018dnnbuilder}, every DNN layer has its own accelerator without resource sharing.
    The implementation target is usually throughput.
    The $Perf_{loss}$ computation follows Eq.~\ref{eq:Lat_max}, and the
    $RES$ computation follows Eq.~\ref{eq:res_throughput}.
    
    }
\end{itemize}

Either way, the operation performance $Perf^q(op_i^m)$ will be the operation latency.
We let the operation resource $Res^q(op_i^m)$ be the number of DSPs, which are usually the most critical resource on FPGA.
To formulate $Perf^q(op_i^m)$ and $Res^q(op_i^m)$, we introduce additional implementation variables for FPGA, \textbf{the parallel factors} of the IPs, denoted as $pf_i^m$.
Parallel factors describe the \textit{parallelism}, indicating how many multiplications can be done concurrently.
In FPGA design, since the parallelism usually increases exponentially such as 64, 128, 256, etc., we use the exponential form of $2^{pf}$ to describe parallelism.

\vspace{2pt}
\textbf{4.1.1 Latency --- $Perf^q(op_i^m)$}
\vspace{2pt}

As defined in Section~\ref{sec:nas_design_space}, each operation $op_i^m$ is composed of a set of sequential DNN layers such as convolution, batch normalization and activation, and the latency and resource of $op_i^m$ should be the summation of all layers.
For an operation $op_i^m$ with parallel factor of $pf_i^m$ in $q$-bit,
its latency can be approximated as:
\begin{equation}
    Perf^q(op_i^m) = Lat^q(op_i^m) = \sum_{l\in op} (lat_l)
    \vspace{-2pt}
\end{equation}
\begin{equation}
    lat_l = \Phi(q) \times \begin{cases}
   2^{-pf_i^m} \cdot k^2 \cdot h \cdot w \cdot c^{in} \cdot c^{out} & \mbox{if } l \mbox{ is conv.} \\
   2^{-pf_i^m} \cdot k^2 \cdot h \cdot w \cdot c^{in} & \mbox{if } l \mbox{ is dw-conv.} \\
   2^{-pf_i^m} \cdot h \cdot w \cdot c^{in} & \mbox{otherwise} \\
\end{cases}
\label{eq:Lat_op_cases}
\vspace{-2pt}
\end{equation}


In the above equation, $k$ is the convolution kernel size;
$h$, $w$, $c^{in}$ and $c^{out}$ represent the data dimension of operation $op_i^m$;
$\Phi(q)$ is the calibration for latency under bit-width of $q$.
Intuitively, smaller bit-width leads to shorter latency because of less off-chip data movement and less computation.
For simplification, we let $\Phi(q)=q$ to simulate such a phenomenon.

\vspace{2pt}
\textbf{4.1.2 Resource --- $Res^q(op_i^m)$}
\vspace{2pt}

The resource (number of DSPs) of the IP with parallel factor of $pf_i^m$ in $q$-bit can be approximated as:
\begin{equation}
    Res^q(op_i^m) = \Psi(q) \times  2^{pf_i^m} 
\label{eq:R_op_cases}
\end{equation}
where $\Psi(q)$ is the calibration for resource under bit-width of $q$. 
On FPGA, the number of DSPs is non-linear to bit-width. 
For example, if the data precision is lower than 8-bit, then two multiplications can be calculated on one Xilinx DSP48, reducing the DSP usage by half;
if the data precision is lower than 4-bit, we assume that multiplications are computed using Lookup-tables (LUTs).
Therefore, we use a piece-wise function to describe $\Psi(q)$:
$\Psi(q)=1$ when $9 \leq q \leq 16$;
$\Psi(q)=\frac{1}{2}$ when $5 \leq q \leq 8$;
$\Psi(q)=0$ when $q \leq 4$.


\vspace{-4pt}
\subsection{GPU\label{sec:gpu}}
On GPUs, the most widely used performance metric is latency, so we let $Perf_{loss}$ to be the latency assuming the batch size is 1.
We assume the resource is fixed given a GPU.
Since GPU latency is relatively easy to measure, we use normalized latency from directly measured values to represent inference latency under $q$-bit data precision.
Therefore, $Perf^q(op_i^m)$ is a constant under a specific $q$.
Currently, the GPU data precision is greatly restricted by the framework support.
Since the current TensorRT only supports 8-bit fixed and 16-/32-bit floating data, we limit data precisions to be 8/16/32-bit for now but can be easily extended to support more bitwidths.
Meanwhile, since the current mixed precision inference has not been well supported by GPU development framework, we constrain the overall DNN to use the same data precision.
Therefore, $\forall i, m$, we have $\phi_{i, m, q} = \phi_q$, which simplifies Eq.~\ref{eq:imp_op_i_m}.

\vspace{-8pt}
\subsection{Dedicated Accelerators}

Besides GPU and FPGA, there are also dedicated ASIC accelerators for efficient DNN implementation, such as
Stripes \cite{judd2016stripes}, Loom \cite{sharify2018loom} and Bit-Fusion \cite{sharma2018bit}, which are DNN accelerators that support dynamic data precisions efficiently.
As an example, in the Loom \cite{sharify2018loom} work, the computation latency and energy of convolution layers scale inversely and almost  proportionally with the precisions of weights and activations.
Our proposed method can be directly applied to such accelerators as well, by formulating the latency and energy of an operation $op_m^i$ proportionally to data precision. We will leave this for future work.


\vspace{-6pt}
\section{overall algorithm} 


First of all, the DNN variables and implementation variables are initialized, including the number of blocks $N$, the number of operation candidates $M$, and sampling possibilities $\Theta$ and $\Phi$.
For recursive FPGA architecture, the algorithm initializes $M$ IP instances, each with an initial parallel factor $pf_0=log(\frac{RES_{ub}}{M})$.
For pipelined FPGA architecture, the algorithm initializes $M \times N$ parallel factors for all the operation candidates, each $pf_0=log(\frac{RES_{ub}}{M\times N})$.
For different devices, only the initializations are different, and the remaining co-search follows the same procedure.

After initialization, the algorithm optimizes Eq.~\ref{eq:co_search_obj_func} using stochastic gradient descent on the fused search variables $\{A, I\}$, following a bilevel approach~\cite{liu2018darts}.
In each iteration, it first fixes $\Theta$, $\Phi$ and $I$, and updates DNN weights $\omega$ by minimizing the training loss on training dataset.
Then, it fixes the DNN weights $\omega$ and updates $\Theta$, $\Phi$ and $I$ by descending Eq.~\ref{eq:co_search_obj_func} on the validation set.
The algorithm iterates until DNN training converges or reaches a fixed number of epochs.
Finally, the searched DNN needs to be trained from scratch on the target dataset, e.g. ImageNet, and the implementation variables, such as parallel factors, also need to be re-tuned.

\begin{figure*}
\vspace{-18pt}
    \centering
    \includegraphics[width=0.92\textwidth]{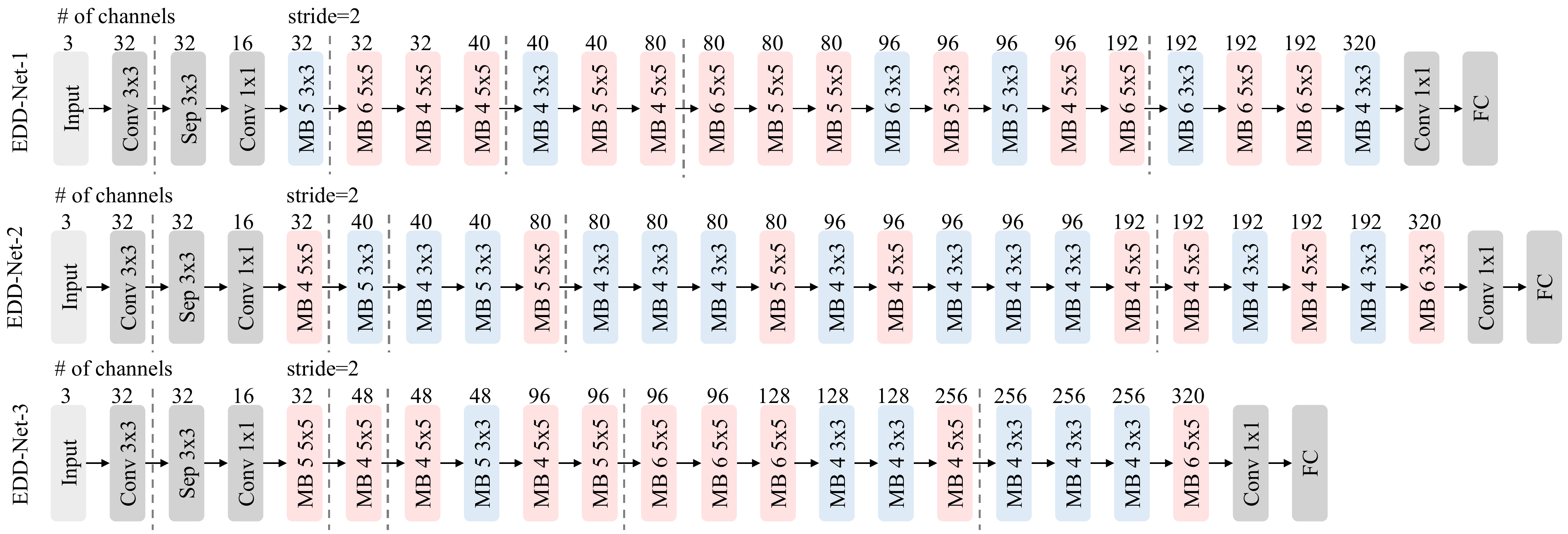}
    \vspace{-10pt}
    \caption{Architectures of three EDD-Net models. EDD-Net-1 targets GPU, EDD-Net-2 targets recursive FPGA accelerator, and EDD-Net-3 targets pipelined FPGA accelerator.}
    \label{fig:DNN_results}
    \vspace{-12pt}
\end{figure*}

\begin{table}[t]
    \centering
    \renewcommand{\arraystretch}{0.90}
    \small
    \caption{Comparisons with existing NAS solutions}
    \label{tab:nas_comp}
    \vspace{-8pt}
    \setlength{\tabcolsep}{2pt}
    \begin{tabular}{c|c|c|c|c}
    \hline
    & \multicolumn{2}{c|}{Test Error (\%)} & GPU Latency & FPGA Latency\\ 
    \hline
    & Top-1 & Top-5 & Titan RTX & ZCU102 \cite{CHaiDNN}\\
    \hline
    \multicolumn{5}{l}{\textbf{Baseline Models}}\\
    \hline
    GoogleNet  & 30.22 & 10.47 & 27.75 ms & 13.25 ms \\
    MobileNet-V2 \cite{sandler2018mobilenetv2}  & 28.1 & 9.7 & 17.87 ms & 10.85 ms\\
    ShuffleNet-V2 \cite{ma2018shufflenet} & 30.6 & 11.7 & 21.91 ms & NA \\
    ResNet18  & 30.2 & 10.9 & 9.71 ms & 10.15ms \\
    \hline
    \multicolumn{5}{l}{\textbf{Hardware-aware NAS Models}}\\
    \hline
    MNasNet-A1 \cite{tan2019mnasnet}  & 24.8 & 7.5 & 17.94 ms & 8.78 ms\\
    FBNet-C \cite{wu2019fbnet} & 24.9 & 7.6 & 22.54 ms & 12.21 ms\\
    Proxyless-cpu \cite{cai2018proxylessnas} & 24.7 & 7.6 & 21.34 ms & 10.81 ms \\
    Proxyless-Mobile \cite{cai2018proxylessnas} & 25.4 & 7.8 & 21.23 ms & 10.78 ms \\
    Proxyless-gpu \cite{cai2018proxylessnas} & 24.9 & 7.5 & 15.72 ms & 10.79 ms \\
    \hline
    \textbf{EDD-Net-1 } & 25.3 & 7.7 & \textbf{11.17 ms} & 11.15 ms \\
    \textbf{EDD-Net-2 } & 25.4 & 7.9 & 13.00 ms & \textbf{7.96 ms} \\

    \hline
    \end{tabular}
    \\
    \vspace{-4pt}
\end{table}

\begin{table}[t]

    \centering
    \renewcommand{\arraystretch}{0.9}
    \small
    \caption{EDD-Net-1 accuracy and latency on 1080 Ti}
    \label{tab:edd-net-1}
    \vspace{-8pt}
    \setlength{\tabcolsep}{2pt}
    \begin{tabular}{c|c|c |c}
    \hline
    & 32-bit Floating & 16-bit Floating & 8-bit Integer \\\hline
    Test Error & 25.5\% & 25.3\% & 26.4\% \\ \hline
    Latency & 2.83 ms &  2.29 ms & 1.74 ms \\

    \hline
    \end{tabular}
    \vspace{-4pt}
\end{table}

\begin{table}[t]

    \centering
    \renewcommand{\arraystretch}{0.9}
    \small
    \caption{Comparison of EDD-Net-3 with DNNBuilder\cite{zhang2018dnnbuilder}}
    \label{tab:edd-net-3}
    \vspace{-8pt}
    \setlength{\tabcolsep}{2pt}
    \begin{tabular}{c|c|c |c}
    \hline
     & Top-1 Error (\%)  & Top-5 Error (\%) & Throughput (ZC706) \\\hline
    VGG16 & 29.5 & 10.0 & 27.7 fps \\\hline
    EDD-Net-3 & 25.6 & 7.7 & 40.2 fps\\

    \hline
    \end{tabular}
    \vspace{-4pt}
\end{table}

\vspace{-8pt}
\section{experiments}

We apply our EDD co-search on a subset of ImageNet dataset randomly sampled from 100 classes.
The searched DNNs are trained from scratch on the entire ImageNet with 1000 classes.
We run for fixed 50 epochs during the EDD search.
The initial DNN has 20 MBConv blocks ($N=20$). Each MBConv has a filter size within $\{3, 5, 7\}$ and a channel expansion ratio within $\{4, 5, 6\}$. So there are $M=3\times3=9$ operations within a MBConv block with different filer sizes and the numbers of channels.
During the search, for GPUs, the DNN weights are 8-/16-/32-bit and activations are 32-bit;
for FPGAs, the DNN weights are 4-/8-/16-bit and activations are 16-bit fixed point.

We demonstrate our EDD methodology targeting three different hardware platforms, each with a searched DNN model, called EDD-Net.
We then compare our DNNs with the ones searched by the state-of-the-art hardware-aware NAS approaches.
The three DNNs are targeting:
(1) low-latency oriented GPU (EDD-Net-1); 
(2) recursive FPGA architecture (EDD-Net-2);
(3) pipelined FPGA architecture (EDD-Net-3).
The three DNNs are shown in Fig.~\ref{fig:DNN_results}; each is produced through EDD within a 12-hour search on a P100 GPU.

First, for GPU-targeted EDD-Net-1, the algorithm suggests the 16-bit precision for weights for the combined objective function including both accuracy and latency.
We compare EDD-Net-1 with the state-of-the-art hardware-aware NAS approaches as shown in Table \ref{tab:nas_comp}, where the GPU inference latency is tested on Titan RTX GPU.
It shows that, EDD-Net-1 reaches the similar accuracy as the state-of-the-art DNN models, while achieving the shortest inference latency, 11.17 ms, which is 1.4$\times$ faster than Proxyless-GPU \cite{cai2018proxylessnas}, the previous best result reported through the NAS approach.
Compared to other mobile-oriented NAS results as a reference, it is 2.0$\times$ faster than FBNet-C \cite{wu2019fbnet} and 1.6$\times$ faster than MNasNet \cite{tan2019mnasnet}.
Table \ref{tab:edd-net-1} shows the accuracy and latency results of EDD-Net-1 on Nvidia 1080 Ti GPU after re-training and fine-tuning using TensorRT under different data precisions.

Second, we intend to compare FPGA-targeted EDD-Net-2 and EDD-Net-3 with existing FPGA/DNN co-design works such as \cite{jiang2019accuracy} and \cite{hao2019fpga}, but neither of them provided accuracy results on ImageNet.
Therefore, to make a relatively fair comparison, for EDD-Net-2, which targets a recursive FPGA accelerator, we adopt the well-recognized CHaiDNN framework~\cite{CHaiDNN}, which is also a recursive FPGA accelerator.
The FPGA latency is collected by running various DNN models with CHaiDNN accelerators under the same data precision on Xilinx ZCU102 FPGA as shown in Table \ref{tab:nas_comp}. The ShuffleNet \cite{ma2018shufflenet} is currently not supported by CHaiDNN.
It shows that EDD-Net-2 on FPGA delivers the shortest latency among the FPGA implementations of all the DNNs, 7.96 ms.
It is 1.37$\times$ faster than the FPGA implementation of ProxylessNet \cite{cai2018proxylessnas}, 1.53$\times$ faster than FBNet \cite{wu2019fbnet} and 1.1$\times$ faster than MNasNet \cite{tan2019mnasnet}.
This result shows that our methodology generalizes well to FPGA and can search for FPGA-friendly DNNs effectively.

Third, EDD-Net-3 is searched targeting a pipelined FPGA accelerator.
In this case,
we limit the total number of DNN blocks because more DNN blocks require more resource and complicated memory control logic.
In Fig \ref{fig:DNN_results}, it shows that EDD-Net-3 is shallower but with more channels and larger kernels.
We compare the throughput of EDD-Net-3 with a state-of-the-art pipelined FPGA accelerator, DNNBuilder~\cite{zhang2018dnnbuilder}, on ZC706 FPGA with 900 DSPs.
As shown in Table \ref{tab:edd-net-3}, under 16-bit fixed point, EDD-Net-3 achieves 1.45$\times$ higher throughput with a much higher accuracy.

\vspace{-8pt}
\section{conclusion}
In this work, we proposed EDD, a fully simultaneous,
efficient differentiable DNN architecture and implementation co-search methodology, which can target different hardware devices with different performance objectives.
We formulated the co-search problem as an elegant differentiable mathematical formulation by fusing DNN architecture search variables and hardware implementation variables into one solution space, considering both accuracy loss and hardware performance loss.
In the experiments, we demonstrated three DNN models, targeting low-latency GPU, recursive FPGA accelerator and pipelined FPGA accelerator, respectively.
Our EDD models deliver similar accuracy as the best existing DNN models searched by NAS on ImageNet, with an improved performance by 1.4$\times$ on GPU and 1.45$\times$ on FPGA.
The future works include GPU power and resource formulation, and EDD search for dedicated accelerators.

\vspace{-8pt}

\subsection*{ACKNOWLEDGEMENT}
This work is supported in part by the IBM-Illinois Center for Cognitive Computing System Research (C3SR), Semiconductor Research Corporation (SRC) and Campus for Research Excellence and Technological Enterprise (CREATE) programme in Singapore.


\bibliographystyle{unsrt}
\footnotesize
\bibliography{ref}

\end{document}